\documentclass[graybox]{svmult}

\usepackage{mathptmx}       % selects Times Roman as basic font
\usepackage{helvet}         % selects Helvetica as sans-serif font
\usepackage{courier}        % selects Courier as typewriter font
\usepackage{type1cm}        % activate if the above 3 fonts are
\usepackage{makeidx}         % allows index generation
\usepackage{graphicx}        % standard LaTeX graphics tool
\usepackage[bottom]{footmisc}% places footnotes at page bottom

\usepackage{amsmath,amssymb,amsfonts}
\newcommand{\R}{\mathbb{R}}
\newcommand{\N}{\mathbb{N}}
\usepackage{subfig,booktabs,array}
\makeindex             % used for the subject index
\begin{document}

\title*{Cracks in concrete}
\author{Tin Barisin and Christian Jung and Anna Nowacka and Claudia Redenbach and Katja Schladitz}
\authorrunning{T. Barisin, C. Jung, A. Nowacka, C. Redenbach, K. Schladitz} 
% for an abbreviated version of
\institute{Tin Barisin \at Fraunhofer ITWM and Technische Universit{\"a}t Kaiserslautern
\and Christian Jung \at Technische Universit{\"a}t Kaiserslautern \email{cjung@mathematik.uni-kl.de}
\and Anna Nowacka \at Fraunhofer ITWM and Technische Universit{\"a}t Kaiserslautern
\and Claudia Redenbach \at Technische Universit{\"a}t Kaiserslautern \email{redenbach@mathematik.uni-kl.de}
\and Katja Schladitz \at Fraunhofer ITWM, Kaiserslautern  \email{katja.schladitz@itwm.fraunhofer.de}
}
\maketitle

% 10--15 lines, teaser, available with unrestricted access, will not appear in the printed version, use * for not including
\abstract{Finding and properly segmenting cracks in images of concrete is a
challenging task. Cracks are thin and rough and being air filled do yield a 
very weak contrast in 3D images obtained by computed tomography. 
Enhancing and segmenting dark lower-dimensional structures is already 
demanding. The heterogeneous concrete matrix and the size of the images 
further increase the complexity. 
ML methods have proven to solve difficult segmentation problems when trained
on enough and well annotated data.  However, so 
far, there is not much 3D image data of cracks available at all, let alone 
annotated. Interactive annotation is error-prone as humans can easily tell 
cats from dogs or roads without from roads with cars but have a hard time 
deciding whether a thin and dark structure seen in a 2D slice continues in 
the next one. Training networks by synthetic, simulated images is an elegant
way out, bears however its own challenges.
In this contribution, we describe how to generate semi-synthetic image data 
to train CNN like the well known 3D U-Net or random forests for segmenting 
cracks in 3D images of concrete. 
The thickness of real cracks varies widely, both, within one crack as well 
as from crack to crack in the same sample. The segmentation method should 
therefore be invariant with respect to scale changes. We introduce the so-called 
RieszNet, designed for exactly this purpose.
Finally, we discuss how to generalize the ML crack segmentation methods to 
other concrete types.
\footnote{This is a preprint of the following chapter: 
Tin Barisin, Christian Jung, Anna Nowacka, Claudia Redenbach, and Katja Schladitz: 
Cracks in concrete, published in 
Statistical Machine Learning for Engineering with Applications (Lecture Notes in Statistics),
edited by Jürgen Franke, Anita Schöbel, 2024, Springer Cham, 
reproduced with permission of Springer Nature Switzerland AG 2024. The final authenticated version is available online at: 
https://doi.org/10.1007/978-3-031-66253-9}
}

\section{Why do we look for cracks in concrete?}
\label{sec:why}

Concrete is brittle material, cracks thus occur naturally. Where and at which load 
they appear, as well as their shape, size, and orientation bear valuable information 
on the particular concrete under investigation. Observations on the surface of 
concrete samples are state of the art. 

Computed tomography (CT) has proven to yield much richer information as it adds the 
third dimension and images crack regions that are sure to not being affected by the 
sample preparation \cite{paetsch12,ansell18}. 
However, wide application in civil engineering research has been hampered by the 
conflict between the high lateral resolution needed to properly observe the cracks 
and the concrete's micro-structure on the one hand and the large size of 
representative samples on the other hand. A dedicated, custom-built CT 
device promises to resolve this conflict \cite{salamon:gulliver}.

\section{Why is it difficult to segment cracks in images of concrete?}
\label{sec:challenge}

\begin{figure}[b]
%\sidecaption
% For example, with the graphicx style use
\centering
\centering
\includegraphics[width=.49\textwidth]{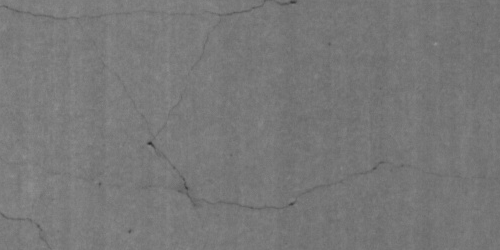}\hfill 
\includegraphics[width=.49\textwidth]{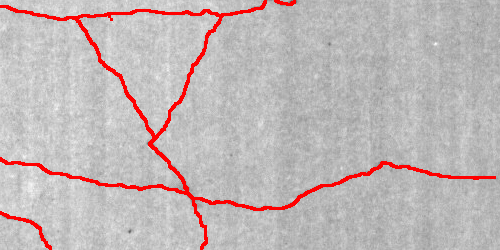}
%
% If no graphics program available, insert a blank space i.e. use
%\picplace{5cm}{2cm} % Give the correct figure height and width in cm
%
\caption{Example of a crack in concrete as they appear in optical images. 
Small cutout from a 2D image of a concrete panel. 600 by 600 pixels cover approximately 
$4.5\,$cm by $4.5\,$cm. Left: original. Right: crack as segmented by CrackNet \cite{mueller18}, see Section~\ref{sec:ml-2d-crack-segmentation}.}
\label{fig:crack-example-2d}       % Give a unique label
\end{figure}

\begin{figure}[b]
%\sidecaption
\includegraphics[width=.48\textwidth]{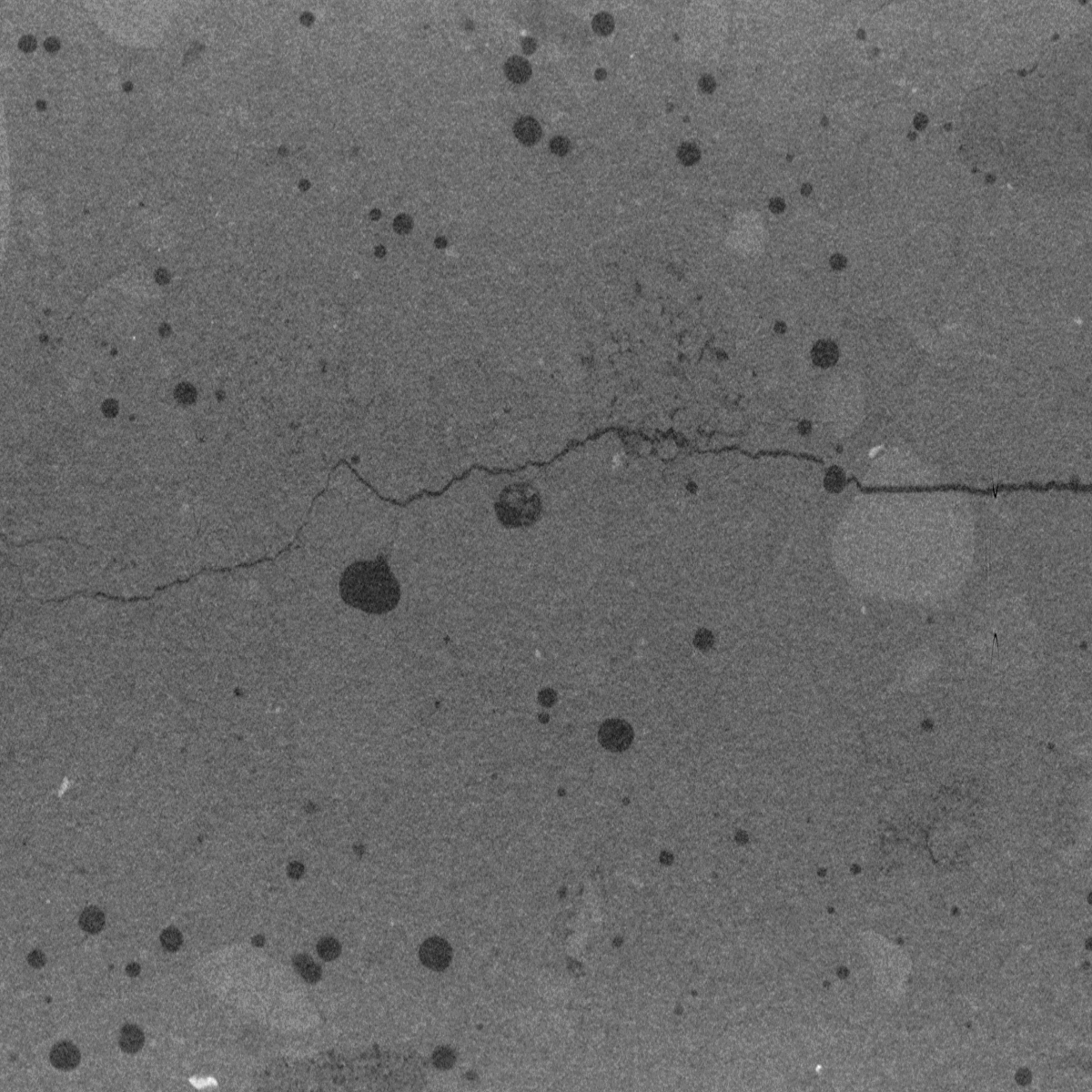}
\includegraphics[height=.48\textwidth,angle=90]{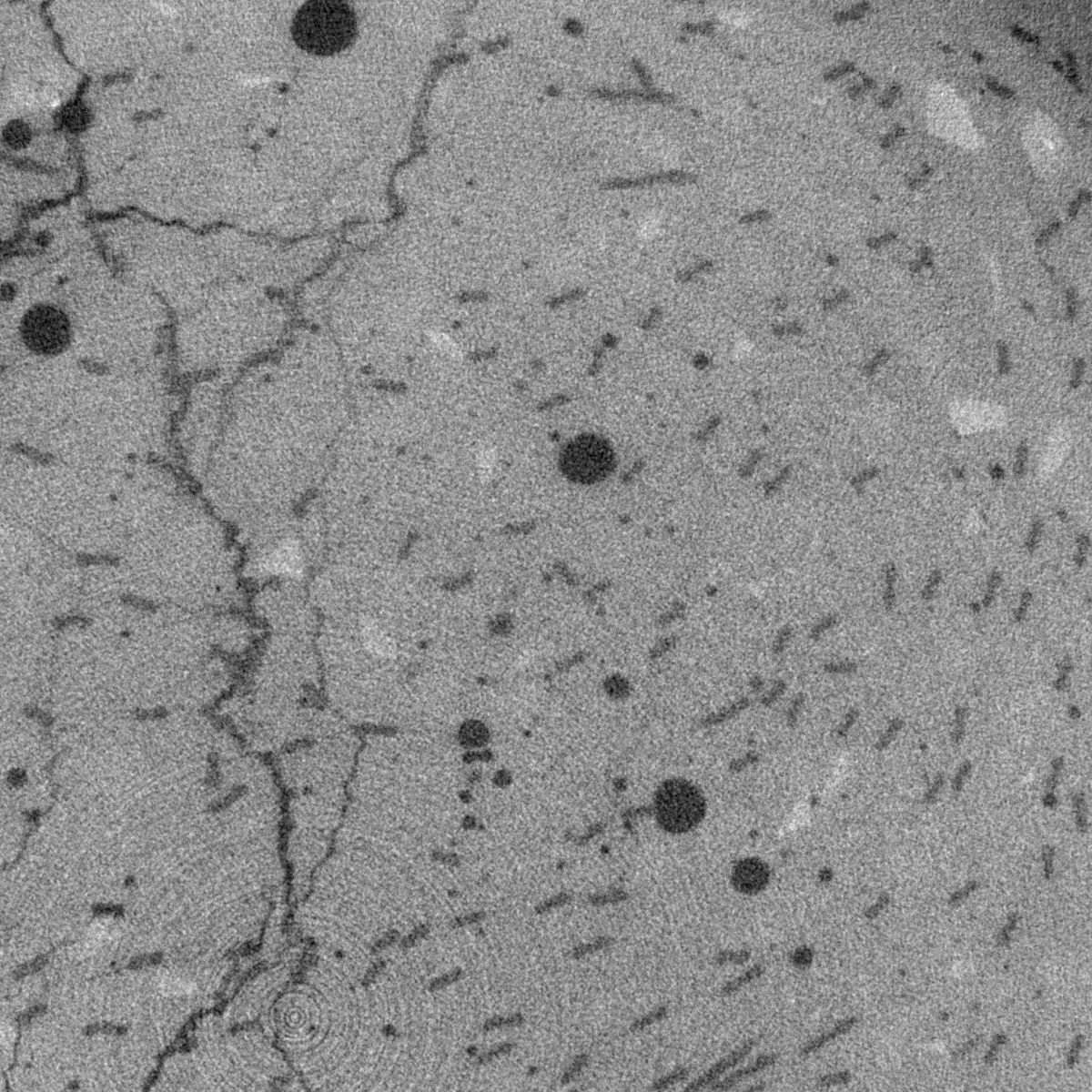}\\
\includegraphics[width=.48\textwidth]{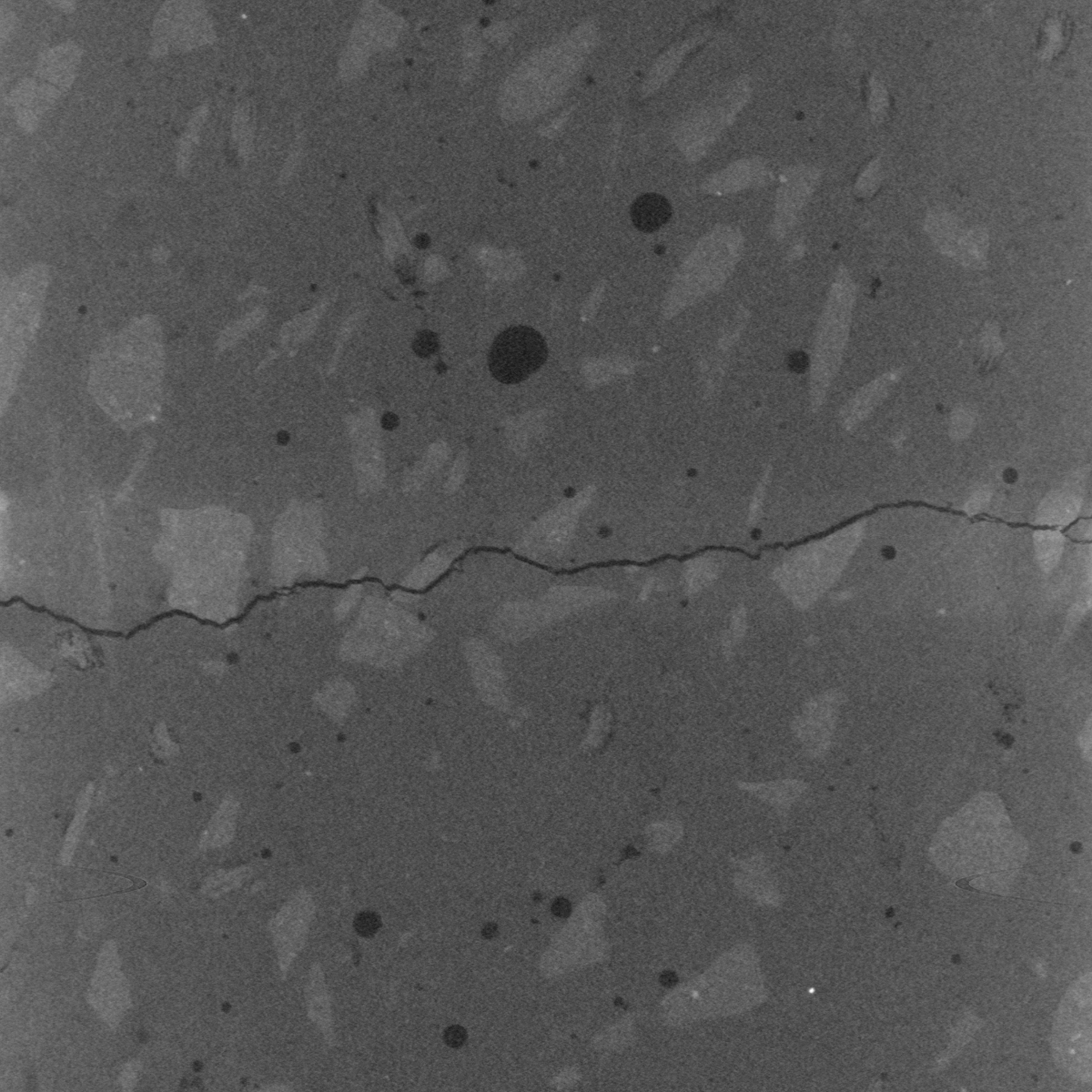}
\includegraphics[width=.48\textwidth]{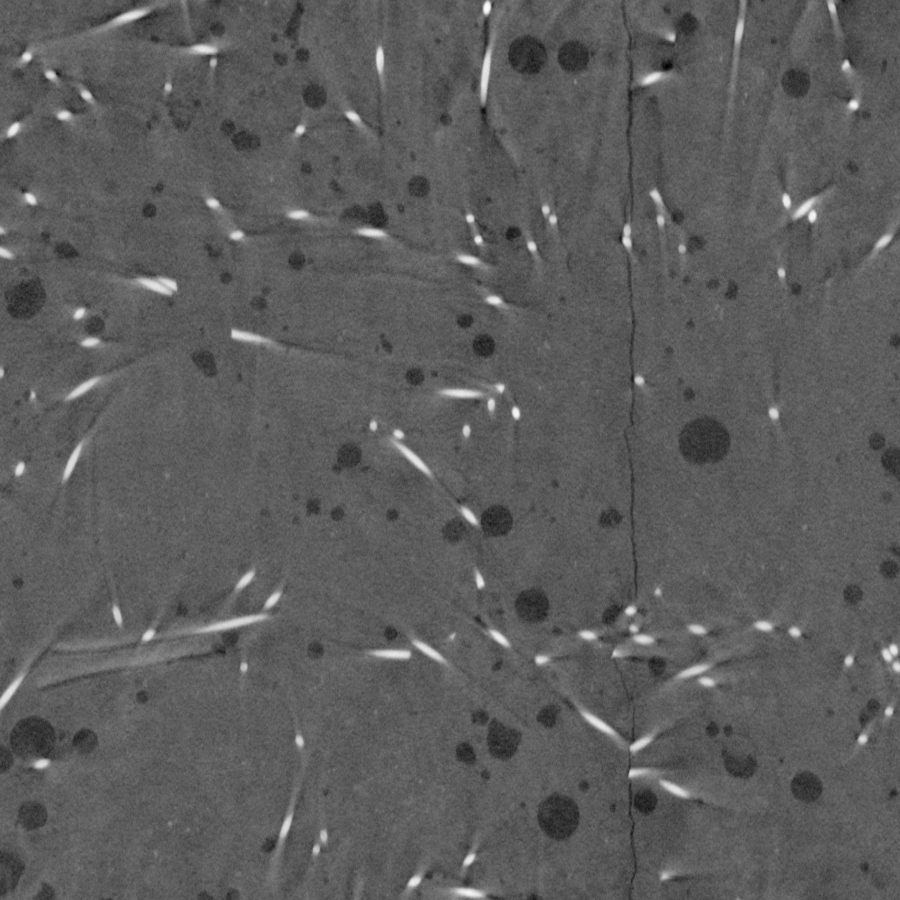}
\caption{Examples of cracks in concrete as they appear in CT images. 
2D slices from the 3D images. All samples created by University of Kaiserslautern, Civil Engineering. 
CT scans by Fraunhofer ITWM, except for top right scanned by Fraunhofer EZRT using the setup \cite{salamon:gulliver}. Left: Normal concrete. Right: Fiber (bottom steel, top polymer) reinforced concrete. Section images contain $1\,200\times1\,200$ pixels covering squares of edge lengths between 2.7 and 4.3 cm.}
\label{fig:crack-examples-3d}       % Give a unique label
\end{figure}

Cracks are thin, essentially rough surfaces in 3D images and paths in 2D images, 
thus lower-dimensional. Cracks are air filled and thus appear dark in images. 
On the usually very heterogeneous concrete structure as background, these are hard 
to perceive even visually, see Figure~\ref{fig:crack-example-2d} for a cutout from 
an industrial optical image and Figure~\ref{fig:crack-examples-3d} for 2D sections 
through reconstructed CT images of several concrete types. In CT images, cracks 
yield a low X-ray absorption contrast. Moreover, concrete contains pores as well. 
Being air filled, too, the pores feature approximately the same gray value 
distribution as the cracks.

Within these noisy gray value images, likely to feature imaging artefacts, we want to 
assign each image pixel or voxel the class crack or the class background. That is, 
in terms of DL, a semantic segmentation problem has to be solved.

DL methods for solving it are additionally challenged by the problem immanent class 
imbalance. Obviously, the vast majority of pixels do not belong to the crack system. 
Hence, besides a lack of training data featuring cracks, the common quality measures
do not differentiate well. 

\section{ML for crack segmentation in 2D images}
\label{sec:ml-2d-crack-segmentation}

In civil engineering, manual marking of cracks on surfaces is still rather the rule 
than an exception. Attempts to automate the task by classical image processing have 
been made \cite{rel:2d-review}, do however either 
not detect the cracks sufficiently reliably or find far too many false positives. 

ML has been used to segment cracks for at least a decade by now \cite{rel:2d-review}. 
CNN approaches are abundant, for example
\cite{ml2-konig21,ml2-shim20,ml2-wang21}, see also the review 
\cite{hsieh-ml-crack-review-20}, 
but random forests and other tree based decision rules are applied successfully, too
\cite{2drf-chun20,2drf-shi16}. All of these methods solve the task given sufficiently large 
and consistent training data. %\todo{Can we estimate how large?} 
Current research focuses on training with as few annotated crack images as possible.  
Semi-supervised \cite{ml2-li20,ml2-shim20,ml2-wang21} or weakly-supervised \cite{ml2-konig21} 
learning is explored to increase robustness. 

In \cite{mueller18}, the well known SegNet \cite{segnet} CNN architecture is 
adapted to segmenting the very fine surface cracks on industrial concrete panels shown in 
Figure~\ref{fig:crack-example-2d}.
In only 353 small subimages of size $224\times224$ pixels, the cracks are marked interactively. 
Augmentation by rotating and flipping the images yields an eightfold larger data set 
consisting of $2\,824$ images. Histogram equalization ensures robustness wrt illumination 
variations. The class imbalance is cared for by weighting the cross-entropy by the 
reciprocals of the class sizes as suggested for SegNet \cite{segnet}. The validation and 
comparison based on 424 images shows that augmentation is crucial for the prediction 
quality. The weighting has a significant yet smaller impact, too. 

\section{Challenge 1: Segment 2D cracks in 3D images}
\label{sec:segment-2d-in-3d}

In \cite{paetsch11}, methods for segmenting cracks in 3D images of concrete obtained by computed 
tomography are reviewed. The so-called Hessian based percolation %from \cite{yamaguchi10} 
is generalized and adapted to 3D. A shape criterion based on the Hessian matrix of 2nd order gray
value derivatives is used to identify dark, locally sheet like structures. These candidate 
regions are allowed to grow as long as the local shape does not change too strongly. For more
details see \cite{paetsch11} and \cite{barisin22}. The latter, more recent comparison includes
3D U-Net and a random forest. Much care is taken to compare the methods in a fair and objective
way. 

Both training ML methods as well as quantitative comparison require data with underlying 
ground truth. In our case, these are 3D gray-value images with known crack structure.
Real CT images with known crack pixels are hard to come by. The only way to generate them 
is interactive annotation of the crack. This is a hard and fast tiring task and it is nearly 
impossible to guarantee spatial consistency when annotating slice wise. 
Semi-synthetic images combining simulated crack structures with the concrete matrix from real CT 
images therefore play a crucial role. 
Two ingredients are needed -- geometric models for the crack 
structure on the one hand and a way to generate the heterogeneous background on the other. We 
discuss two approaches to crack structure modelling as well as how to embed these synthetic 
cracks in realistic CT images in Section~\ref{sec:synthetic-images}.

\subsection{Convolutional neural networks for crack segmentation}
\label{sec:cnn-for-3d-crack-segm}

U-net \cite{unet} is a dedicated CNN architecture for semantic segmentation 
with a well described 3D extension \cite{3dunet}. 
U-net has been used for crack segmentation in 2D images already 
\cite{jenkins-18:unet-for-cracks}. There, and following \cite{jenkins-18:unet-for-cracks} 
also \cite{hsieh-ml-crack-review-20}, the class imbalance is cared for by choosing 
patches including crack pixels with much higher probability during training. 
% We have to  announce that these should be covered. 
% Do we know how many data sets are usually used to train 3d U-nets?
We follow another strategy, closer to the weighting originally suggested \cite{unet}, 
giving crack pixels a higher weight instead of the whole patches. That is:
Let $p_0$ and $p_1$ be the proportion of background (class 0) and crack pixels 
(class 0), respectively. The losses of crack pixels are weighted by 
$w = p_0/p_1>1$ whereas the loss of background pixels keeps weight $w= 1$.

In semantic segmentation, it is common to allow deviations by a couple of 
pixels. Applying such a tolerance is particularly reasonable in a case 
like the cracks where the structure to be segmented is very thin. 

\subsection{Synthetic image data for training crack segmentation CNN}
\label{sec:synthetic-images}

CT imaging is simulated for dimensional measuring where so-called 
traceability is demanded, several software packages are available 
\cite{astra-toolbox15,artist19}, see \cite{artist-etal21} for more. However, so far, the 
results do not reflect the heterogeneity of a typical concrete matrix, where material 
density and chemical composition vary locally very strongly.

Therefore, for the time being, a hybrid technique as described in \cite{barisin22} is 
favored: Simulated cracks are imposed on CT images of crack free concrete samples. 
The gray value distribution in the cracks is derived from the gray value distribution 
in pores. Edges are smoothed slightly to mimic the partial volume effect in CT images:
Gray values in voxels at the boundary of two material components are a weighted mean 
of the gray values of these. As a consequence, voxels at the edge of a crack are 
brighter than those completely within the crack but still darker than concrete matrix 
voxels.

% Thus, here, we describe solely the geometric structures for modelling cracks.
\paragraph{Fractional Brownian motion}
Fractional Brownian surfaces %\cite{brownian-motion-cracks} 
are a model for crack structures 
readily available as a MATLAB function \cite{brownian-surface-matlab}. Its only parameter -- 
the Hurst index $H$ -- controls the roughness of the surfaces. The index takes values in 
$[0,1]$ with the surfaces getting smoother with increasing $H$. Finally, the output of \cite{brownian-surface-matlab} is discretized to yield a binary (black and white) image of the crack, see \cite{barisin22} for details.

\begin{figure}[b]
%\sidecaption
\centering
\includegraphics[width=.4\textwidth]{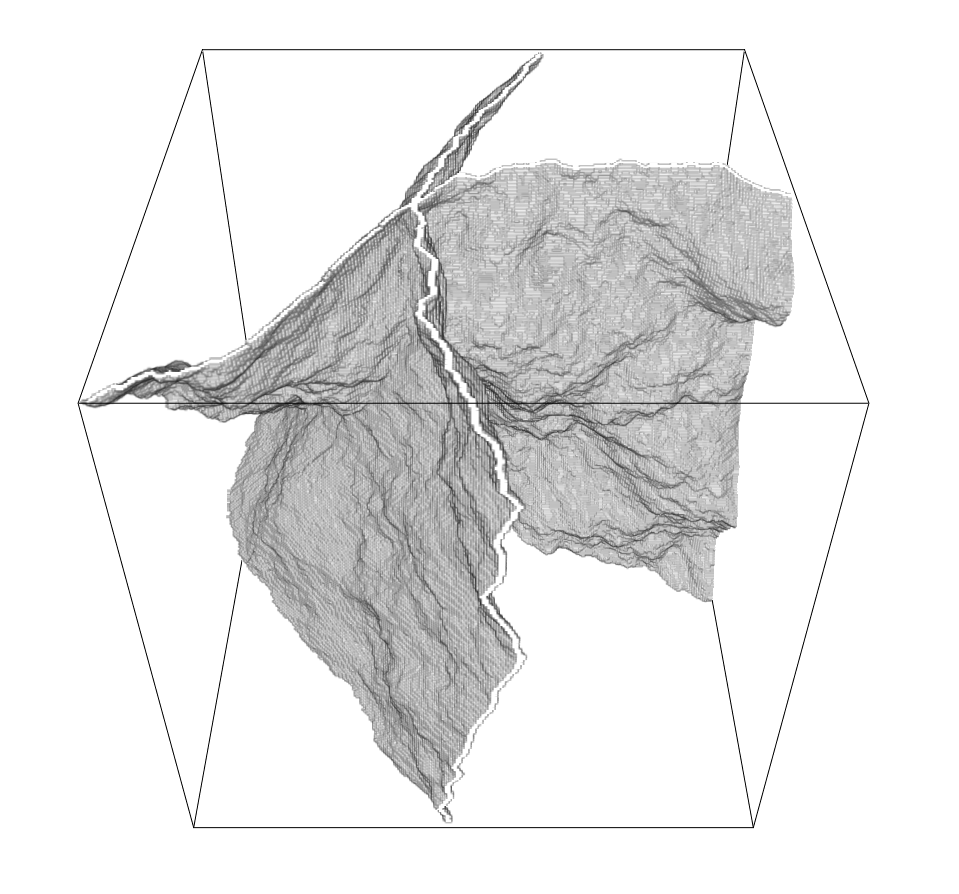}
\includegraphics[width=.42\textwidth]{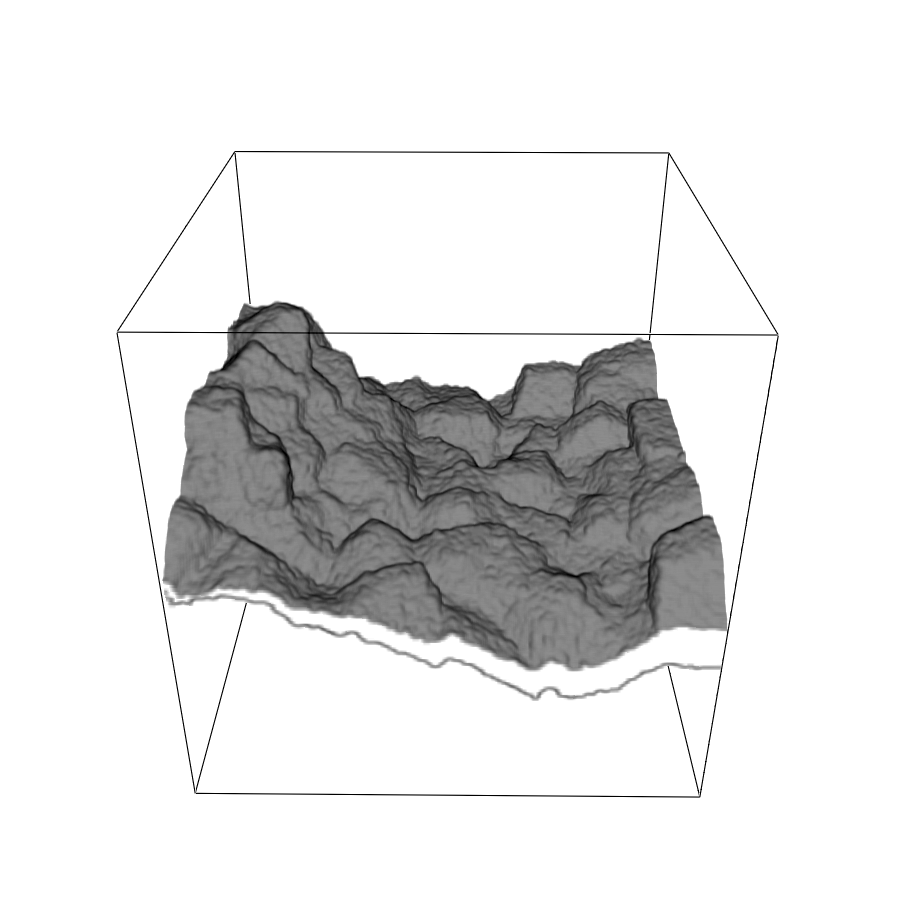}\\
\includegraphics[width=.4\textwidth]{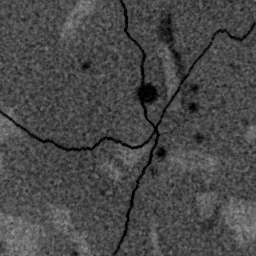}\ 
\includegraphics[width=.4\textwidth]{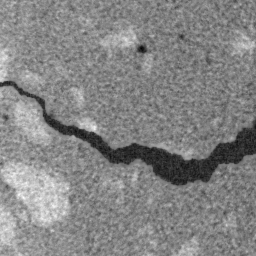}
\caption{Examples of simulated cracks in $256\times256\times256$ voxel images. 
Left: two fractional Brownian surfaces, both widened to constant thickness 3 pixels and with Hurst index $0.97$. 
Right: a minimal surface from a spatial Voronoi tessellation, width varying according to a Bernoulli random walk with parameter. 
Top: volume renderings. 
Bottom: 2D slices of the 3D images of the cracks superimposed on CT images of high performance concrete, scanned with a voxel size of $23.5\,$\textmu m. The visualized cubes and slices have therefore edge length $6\,$mm.
}
\label{fig:crack-examples-synthetic}       % Give a unique label
\end{figure}

Combining realizations allows for crossing cracks. An example is shown on the left side of Figure~\ref{fig:crack-examples-synthetic}. The local 
thickness can be altered by superposition with a suitable function. Superimposed on CT images of concrete, 
they yield image data for both purposes -- training and validation, as shown in \cite{barisin22}.
However, several features of real cracks are hard to capture like
branching, variation of roughness within one crack, and sudden changes of direction. 

We generate $60$ images of size $256^3$ voxels with synthetic cracks of constant thickness $1$, $3$, and $5$ voxel edge lengths, $20$ images of each. 

\paragraph{Minimum weight surfaces in 3d Voronoi diagrams}
Therefore, a new, more versatile model has been suggested recently \cite{jung22cracks}. 
A minimal surface is extracted from a realization of a random spatial 
Voronoi tessellation. As in the case of the fractional 
Brownian surfaces, the stochastic nature of the model enables fast 
generation of as many crack surfaces as needed and naturally captures 
their heterogeneity.

In its simplest version, generated by a homogeneous Poisson point process, 
just one parameter drives the tessellation, namely the intensity of the 
point system. The denser the point system, the smaller the 
Voronoi cells and consequently also the facets finally forming the crack surface.

Given the realization of the tessellation, assign weights to all arcs and facets by 
respective functions. Choose one vertex on each of the vertical edges of the cuboidal window.
Connect them by shortest paths along the faces of the window using Dijk-
stra’s algorithm. A minimum-weight surface bounded by the cycle formed by these paths
can be found by solving an integer program \cite{jung22cracks}.

Modelling the crack by a system of facets opens a variety of additional 
degrees of freedom by marking and nesting. Moreover, the crack surface 
being a subsystem of a spatial tessellation, 2D or 1D branches can be 
generated rather naturally.  

\paragraph{Locally varying thickness by adaptive dilation}
The locally varying thickness featured by real cracks can be captured by 
adaptive dilation of the crack. That is, a morphological dilation with 
size of the structuring element varying according to a size map is applied.
Currently, in each 2D $x$-slice of the 3D image of the crack, the 
foreground is dilated $N_i$ times by a 2 by 2 pixel square structuring 
element, where $i$ is the slice index. The $N_i, \ i=0,\dots,n$ are 
incremented following a 
Bernoulli distribution with parameter $0<p<1$. That is, the $N_i$ 
increase with $i$ and $N_i$ is binomially distributed with parameters 
$(i,p)$. See Figure~\ref{fig:crack-examples-synthetic}, right side for an example.

\subsection{Results: 3D U-Net on synthetic cracks}
\label{sec:res-unet-synth}
We use a 3D U-Net with exactly the original architecture \cite{3dunet} and
the same activation function. As in \cite{unet}, a dropout layer 
at the bottleneck between decoder and encoder helps to prevent overfitting. 
The dropout probability is $0.5$. The learning rate decays by $0.5$ after every 
fifth epoch. As in the 2D case, we weight the binary cross entropy loss by the 
reciprocals of the class sizes in order to cope with the class imbalance. 
%That is, a crack voxel's loss is weighted by $\#$concrete voxels/$\#$ crack voxels. 

We split the $60$ images described above into $9$ for training, $3$ for validation, 
and $48$ for evaluation. The $256^3$ voxel images are divided into patches of 
$64^3$ voxels overlapping by $14$ voxels to avoid edge effects.
Augmentation by rigid motions, cropping and zooming, sharpening and blurring, as well 
as gray value distortions and gray value variations triples the number of 
patches yielding $768$ for training. Gray values are normalized.

The thus trained 3D U-Net segments the synthetic cracks very well, keeping 
mean recall and precision above $0.9$ even for crack width $1$ and no 
tolerance. In particular for these thin cracks it thereby performs better 
than all classical competitors \cite{barisin22}.

\section{Challenge 2: Segment real cracks}
\label{sec:real-cracks}
The most striking difference of real cracks compared to the fixed width 
synthetic cracks considered so far is their widely varying thickness. 
A range of $1-20$ is not at all unusual. The example featured in 
Table~\ref{tab:figs-results-multiscale-real} show that 3D U-Nets trained solely on 
fixed width cracks fail to generalize to these multi-scale cracks. 

\subsection{Multiscale 3D U-Net}
\label{sec:unet-multiscale}
Image pyramids are a fast and straightforward approach to processing 
multi-scale structures. This image representation is based on 
downsampling the image size by fixed factors, here $2$.  
The new voxel gray values are interpolated, here by splines of order $3$. 
Iteration yields a sequence of images starting with the original one and 
getting coarser with each step. We use just three levels, the original 
image, one with doubled, and one with quadruplicated voxel edge length.

Downsampling shrinks thicker cracks to thinner ones in the number of voxel 
edge lengths. A crack width $20$ on the original level is way out of the 
range the U-net has been trained on. However, this original crack width
corresponds to $5$ after downsampling twice by factor $2$.  Hence, 
thin cracks can be segmented at the lower levels, thicker cracks at the 
higher levels of the image pyramid. 

Our multiscale 3D U-Net consists of a 3D U-Net on each level of the 
pyramid. The result is upsampled to the original image size. The 
results of all levels are fused by a simple voxelwise maximum. More 
precisely, the same threshold is applied on each level separately and the 
voxelwise maximum is applied to the resulting binary images. 
Consequently, a voxel is segmented a crack voxel as soon as it is 
found to be a crack voxel on one pyramid level. 
This strategy has proven to work \cite{ict22}, is however costly. 

\subsection{3D U-Net fine-tuned using multiscale cracks}
\label{sec:unet-finetuned-multiscale}
The synthetic Voronoi tessellation based multiscale cracks from 
Section~\ref{sec:synthetic-images} 
enable an alternative strategy, closer to a natural learning process: 
We take our 3D U-Net trained on the images featuring constant width cracks 
from \ref{sec:res-unet-synth} and further train it on five of 
these multiscale crack images.

\subsection{A dedicated network: the Riesz network}
\label{sec:riesz-network}
A far more elegant solution is to impose scale invariance on the neural 
network as devised by Barisin \cite{barisin22riesz}. The main idea is to 
replace the convolutions in the encoder by Riesz transforms whose scale 
equivariance is known and has been exploited in image processing before 
\cite{riesz-paper2019,unser10}.

First and second order Riesz transforms have proven to be useful as 
low level features in %measuring similarity \cite{zhang10}, 
analyzing 
and classifying textures \cite{depeursinge12}, and estimating
orientations \cite{reinhardt20} in 2D images. 
Unser \cite{unser10} 
built steerable wavelet frames, so-called Riesz-Laplace wavelets based on 
the Riesz transform. These are the first ones utilizing the scale 
invariance property of the Riesz transform, and have inspired the 
quasi monogenic shearlets \cite{hauser14}. Recently, Riesz transforms 
have been applied in DL as supplementary features to improve robustness 
\cite{joyseeree19}.

The first order Riesz transform 
$\mathcal{R}=(\mathcal{R}_1,\cdots,\mathcal{R}_d)$ of a 
$d$-dimensional image $f:\R^d \to \R$ is defined as
 \begin{equation*}
     \mathcal{R}_{j}(f)(x) = C_d \lim_{\epsilon \to 0}{\int_{\R^d \setminus B_{\epsilon}}{\frac{y_jf(x-y)}{|y|^{d+1}}dy}}, 
 \end{equation*}
where $C_d = \Gamma((d+1)/2)/\pi^{(d+1)/2}$ is a normalizing constant 
and $B_{\epsilon}$ the ball of radius $\epsilon$ centered at the origin.  
Higher-order Riesz transforms are derived  as sequences of first order Riesz transforms: 
$$ \mathcal{R}^{(j_1,j_2,...,j_d)} (f)(x) := \mathcal{R}_{j_1}( \mathcal{R}_{j_2}( \cdots (\mathcal{R}_{j_d}(f(x))))$$  
for $j_1,j_2,...,j_d\in \{1,\cdots,d\}$.

The Riesz transform has several desirable properties like translation 
invariance or an all-pass filter property \cite{unser10}. 
Denote by $L_{a}(f)(x) = f(\frac{x}{a})$ a dilation or rescaling operator with
size parameter $a>0$. The scale equivariance of the Riesz transform expresses 
itself as commutation with scaling, formally
$$\mathcal{R}_j(L_{a}(f))(x) = L_{a}(\mathcal{R}_j(f))(x).$$

The base layer of a Riesz network a linear combination of first and second order 
Riesz transforms of several orders implemented as 1d convolution across feature 
channels. Thus, in the $K$th layer, the $i$th input channel is linked to the $j$th output channel by the linear combination 
\begin{equation}
    J^{(j,i)}_K(f) = C_0^{(j,i,K)} + \sum_{k=1}^{d}{C_k^{(j,i,K)} \mathcal{R}_k(f)} + \sum_{k=1}^{d}\sum_{l=k}^{d}{C_{k,l}^{(j,i,K)} \mathcal{R}^{(k,l)}(f)},
\end{equation}
where the parameters 
$\{C_0^{(j,K)}, C_k^{(j,i,K)}, C_{k,\ell}^{(j,i,K)} \vert k\in\{1,\cdots,d\}, \ell>k \}$ 
are learned during training. Assume now that the $K$th network layer takes input 
\[F^{(K)} = \left(F^{(K)}_1,\cdots,  F^{(K)}_{c^{(K)}}\right)\in \R^{H\times W \times c^{(K)}}\]
with $c^{(K)}$ feature channels and generates output  
\[F^{(K+1)}=\left(F^{(K+1)}_1,\cdots,F^{(K+1)}_{c^{(K+1)}}\right)\in \R^{H\times W \times c^{(K+1)}}\]  
with $c^{(K+1)}$ channels. Then the output in channel 
$j \in \{1,\cdots,c^{(K+1)} \}$ is
\begin{equation}
    F^{(K+1)}_j = \sum_{i=1}^{c^{(K)}} J_{K}^{(j,i)}\left(F^{(K)}_i\right).
    \label{full:layer}
\end{equation}

The Riesz network is built from layers consisting of the transformations batch 
normalization, Riesz layer, and ReLU (rectified linear unit) activation, 
in this order. Batch normalization improves the training capabilites and avoids 
overfitting, ReLUs introduce non-linearity, and the Riesz layers extract scale 
equivariant spatial features.
A network with $K \in \N$ layers can be simply defined by a $K$-tuple specifying 
the channel sizes e.g. $(c^{(0)}, c^{(1)}, \cdots c^{(K)})$. The final layer is 
a linear combination of the features from the previous layer followed 
by a sigmoid function yielding the desired scale invariant probability map as output.

So far, systematic investigation and comparison of the Riesz network's abilities 
are restricted to the 2d case \cite{barisin22riesz}.  Its extension to 3d is 
nevertheless straightforward. The three layer Riesz network we apply here can 
be written as $(1,16,16,32,1)$. It has 
$(1\cdot 9\cdot16+16)+(16\cdot 9\cdot 16+16)
+(16\cdot 9\cdot 32+32)+(32\cdot1+1)= 7\,153$ trainable parameters. 
This is most remarkable 
when comparing to U-net with more than $2$ million parameters.

\subsection{Results}
We tested the three approaches described above on the remaining 15 images 
(numbered 1 to 15)
featuring Voronoi tessellation based multiscale cracks. We report precision, 
recall, and F1 score for tolerances $0$ and $1$ in 
Table~\ref{tab:results-multiscale}. As had to be 
expected, results for tolerance $2$ are better than those for $1$. 
However no qualitative differences occur. All three methods perform 
worst on image 7 whose gray value distribution  
is indeed strangely distorted, see the original in the first row of
Table~\ref{tab:figs-results-multiscale-synthetic}. It is however 
remarkable, how robust the Riesz network reacts compared to the two 3D U-Nets. 
\begin{table}
\caption{Results of the multiscale and the fine-tuned 3D U-Net and the 
three-layer Riesz network on 15 images with synthetic multiscale cracks for 
tolerances $0$ and $1$. The images yielding the minimal values 
are named in parenthesis.}
\label{tab:results-multiscale}    
\begin{tabular}{|l||r|r|r||r|r|r||r|r|r|}
\hline
\multicolumn{10}{|c|}{tolerance $0$ \rule{0pt}{11pt}}\\
\svhline
method & \multicolumn{3}{|c||}{3D U-Net, multiscale\rule{0pt}{11pt}} 
       & \multicolumn{3}{|c||}{3D U-Net, fine-tuned} 
       & \multicolumn{3}{|c|}{Riesz network} \\
\svhline
measure & \phantom{eda}min & \phantom{di}mean & median & 
          \phantom{eda}min & \phantom{di}mean & median & 
          \phantom{eda}min & \phantom{di}mean & median \rule{0pt}{10pt}\\
\svhline
precision & $0.822$ (10) & $0.920$ &  $0.931$ 
          & $0.880$ (10) & $0.976$ &  $0.984$ 
          & $0.671$ (10) & $0.848$ &  $0.854$ \\
recall & $0.182$ \phantom{1}(7) & $0.859$ & $0.929$ 
       & $0.046$ \phantom{1}(7) & $0.878$ & $0.963$ 
       & $0.805$ \phantom{1}(7) & $0.918$ & $0.920$\\
F1 score  & $0.307$ \phantom{1}(7) & $0.869$ & $0.918$
          & $0.088$ \phantom{1}(7) & $0.897$ & $0.970$
          & $0.773$ (10) & $0.872$ & $0.898$ \\
\svhline
\multicolumn{10}{|c|}{tolerance $1$ \rule{0pt}{11pt}}\\
\svhline
method & \multicolumn{3}{|c||}{3D U-Net, multiscale\rule{0pt}{11pt}} 
       & \multicolumn{3}{|c||}{3D U-Net, fine-tuned} 
       & \multicolumn{3}{|c|}{Riesz network} \\
\svhline
measure & \phantom{eda}min & \phantom{di}mean & median & 
          \phantom{eda}min & \phantom{di}mean & median & 
          \phantom{eda}min & \phantom{di}mean & median \rule{0pt}{10pt}\\
\svhline
precision & $0.962$ (12) & $0.989$ & $0.992$
          & $0.991$ \phantom{1}(4) & $0.997$ &  $0.998$
          & $0.846$ (10) & $0.923$ & $0.917$\\
recall & $0.374$ \phantom{1}(7) & $0.923$ & $0.984$ 
       & $0.096$ \phantom{1}(7) & $0.913$ & $0.989$
       & $0.942$ \phantom{1}(7) & $0.983$ & $0.985$ \\
F1 score  & $0.544$ \phantom{1}(7) & $0.946$ & $0.986$ 
          & $0.175$ \phantom{1}(7) & $0.930$ & $0.994$  
          & $0.912$ (10) & $0.957$ & $0.961$ \\
\hline
\end{tabular}
%$^a$ Table foot note (with superscript)
\end{table}

\newcommand{\myscale}{.2\textheight}
\newcolumntype{C}{>{\centering\arraybackslash}m{\myscale}}
\begin{table}[htb]
    \caption{Examples of segmented synthetic cracks. 2D slices through the 3D images. Image 7 is the one where the 3D U-Nets seem to fail whereas the Riesz network 
    still reaches considerable accuracy. Image 10 is also a hard one according to the accuracy values in Table~\ref{tab:results-multiscale} yet by far not as difficult as 7.}
    \label{tab:figs-results-multiscale-synthetic}
    \centering
    \begin{tabular}{p{3cm}*2{C}@{}}
    \svhline
     CNN & image 7 & image 10  \\
    \hline
    original & 
    \includegraphics[height=\myscale]{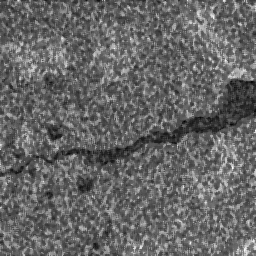} &
    \includegraphics[height=\myscale]{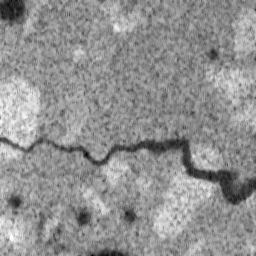} \\
    multiscale 3D U-Net trained on fixed width cracks, 5 voxels (left)  and 1 voxel (right) & 
    \includegraphics[height=\myscale]{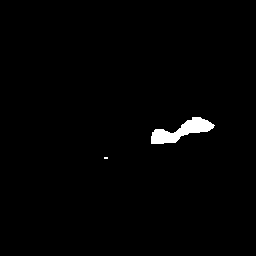} &
    \includegraphics[height=\myscale]{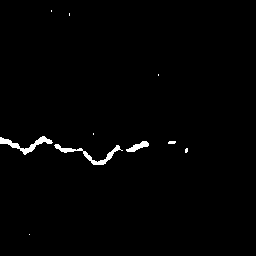}\\
    multiscale 3D U-Net from \ref{sec:unet-multiscale} & \includegraphics[height=\myscale]{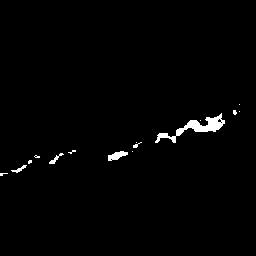} &  \includegraphics[height=\myscale]{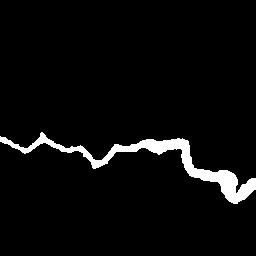}\\
    fine-tuned multiscale 3D U-Net from \ref{sec:unet-finetuned-multiscale} & \includegraphics[height=\myscale]{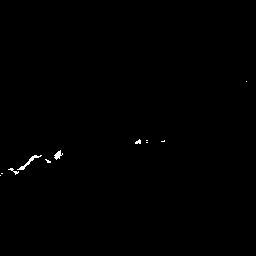} & \includegraphics[height=\myscale]{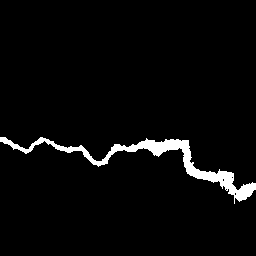}\\  
    Riesz network as described in \ref{sec:riesz-network} & \includegraphics[height=\myscale]{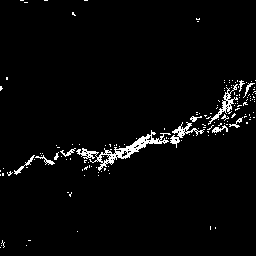} & \includegraphics[height=\myscale]{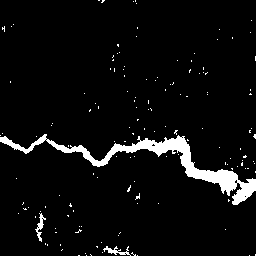}\\
    \svhline
    \end{tabular}
\end{table}

\newcommand{\myscaletwo}{.15\textheight}
\begin{table}[htb]
    \caption{Two real crack systems, segmented by 3D U-Net trained on cracks of 
    thickness $1$, the multiscale 3D U-Net, the fine-tuned 3D U-Net, and the 
    Riesz network. 2D slices through the 3D images.}
    \label{tab:figs-results-multiscale-real}
    \centering
    \begin{tabular}{p{1.5cm}*2{C}@{}}
    \svhline
     CNN & normal concrete & hpc  \\
    \hline
    original & 
    \includegraphics[height=\myscaletwo]{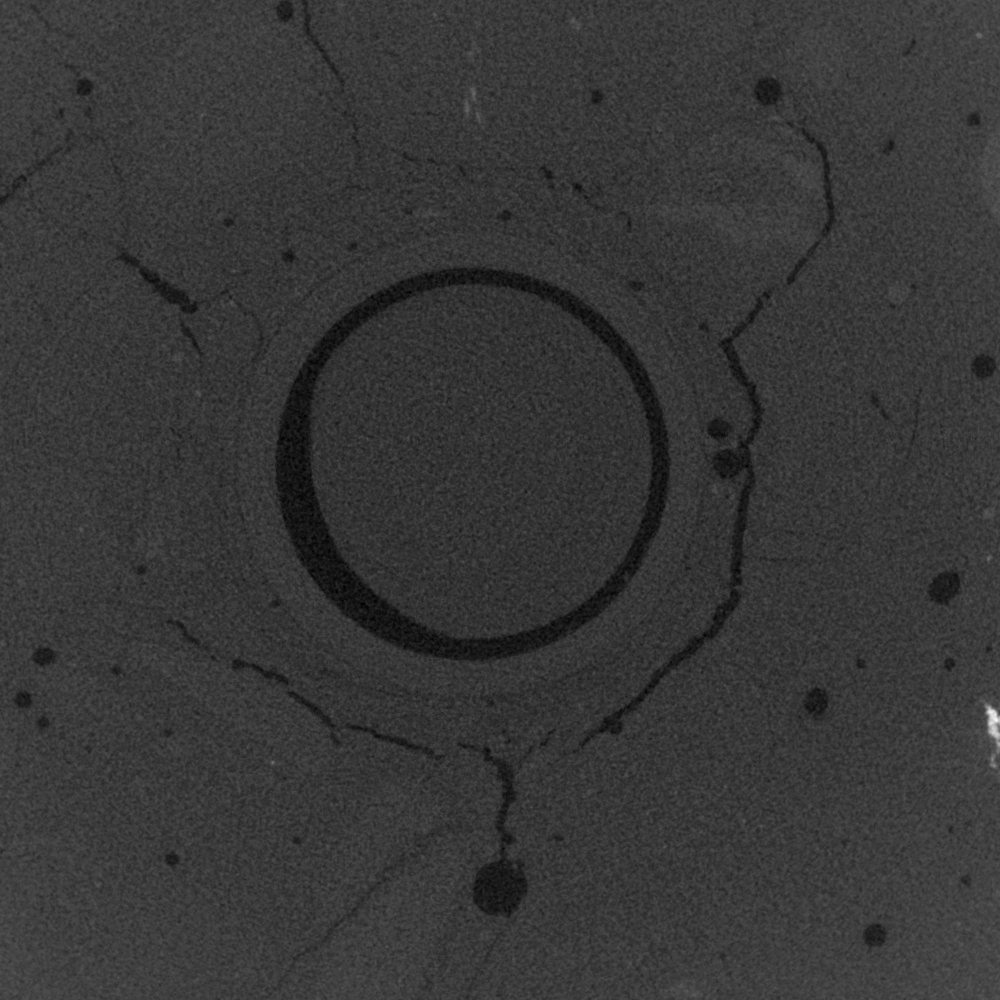} &
    \includegraphics[height=\myscaletwo]{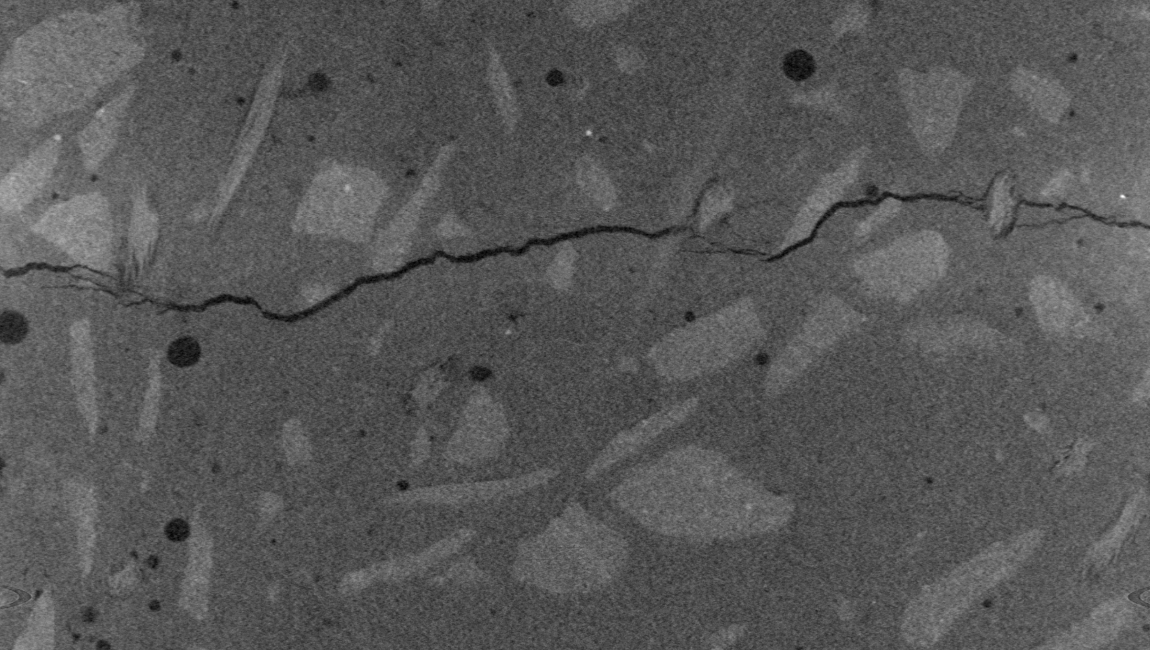} \\
    multiscale 3D U-Net trained on 1 voxel wide cracks & 
    \includegraphics[height=\myscaletwo]{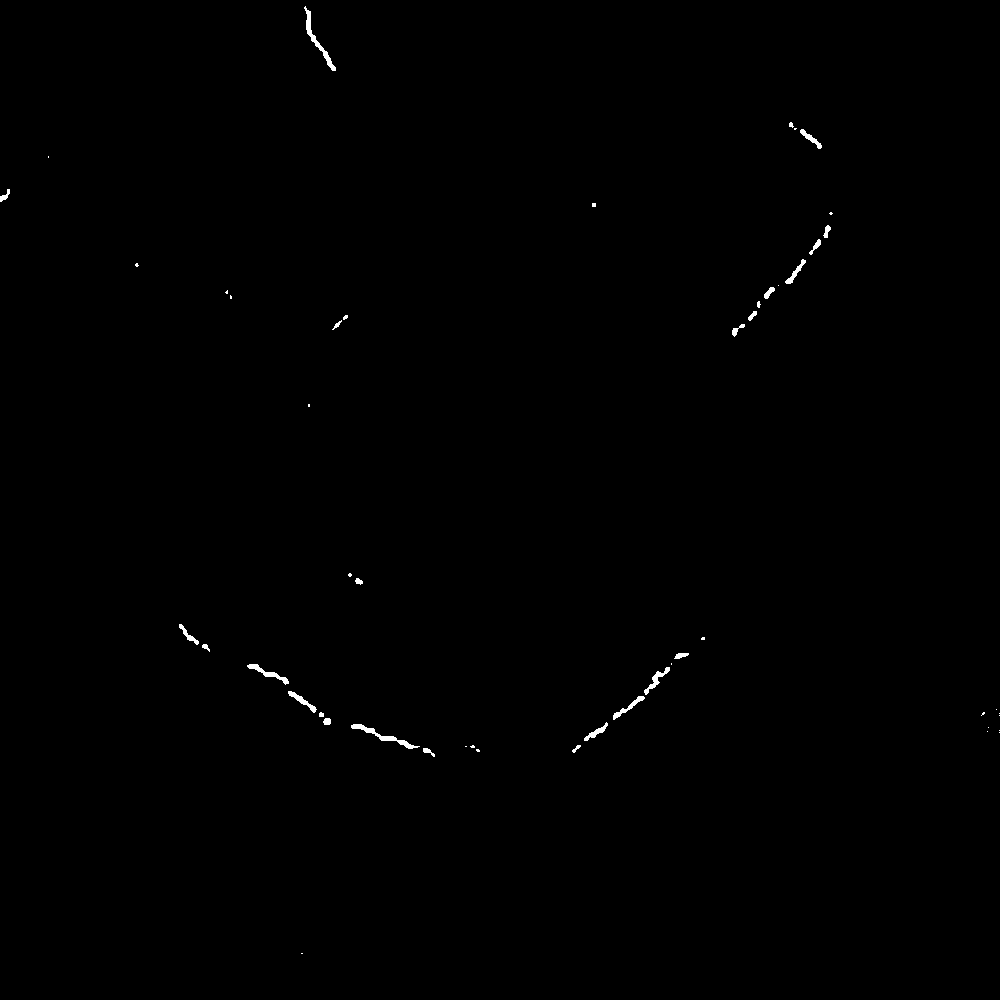} &
    \includegraphics[height=\myscaletwo]{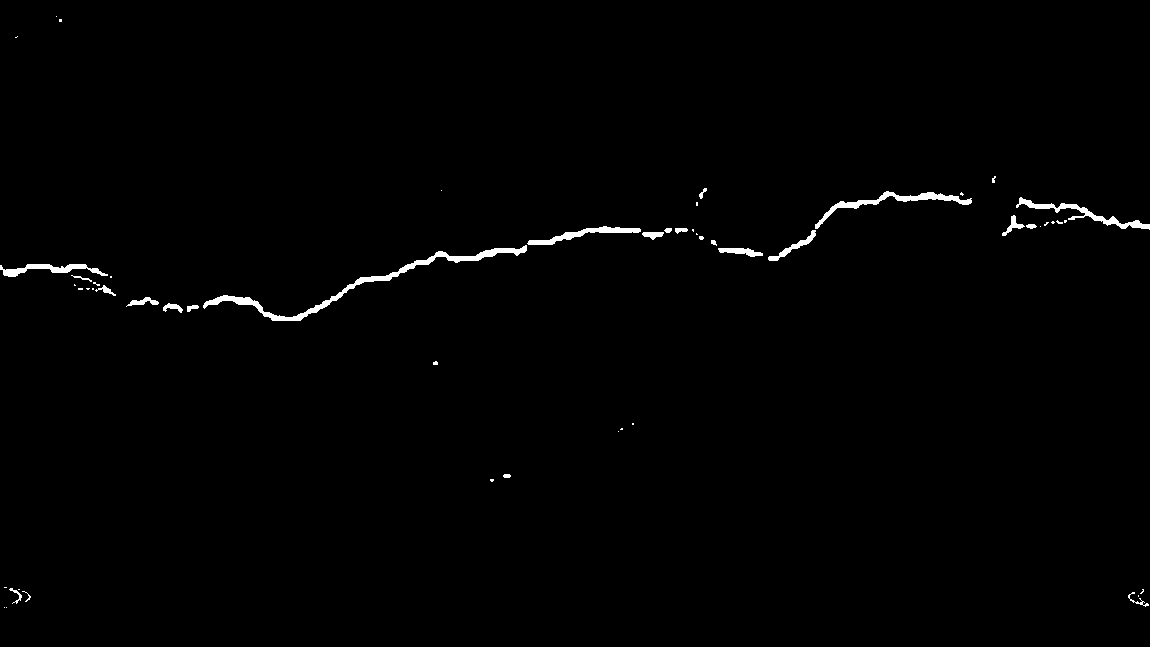}\\
    multiscale 3D U-Net from \ref{sec:unet-multiscale} & \includegraphics[height=\myscaletwo]{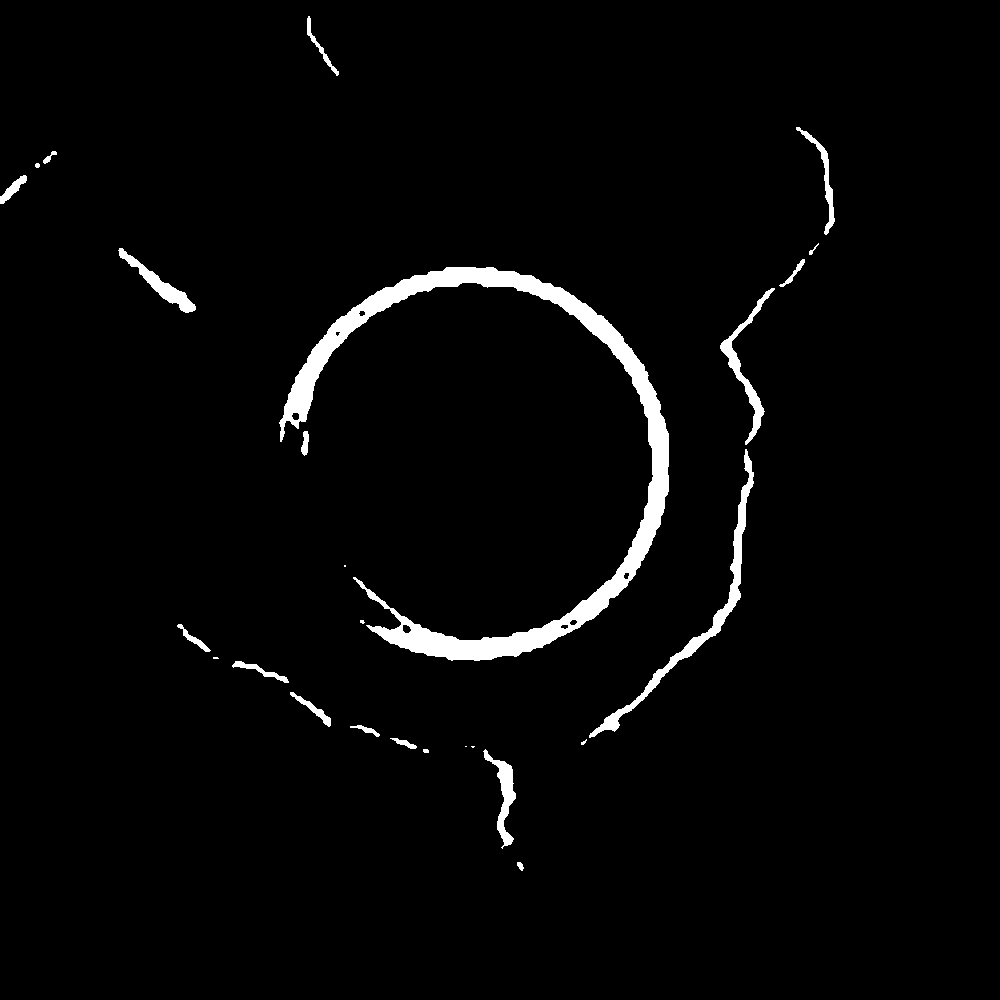} &  \includegraphics[height=\myscaletwo]{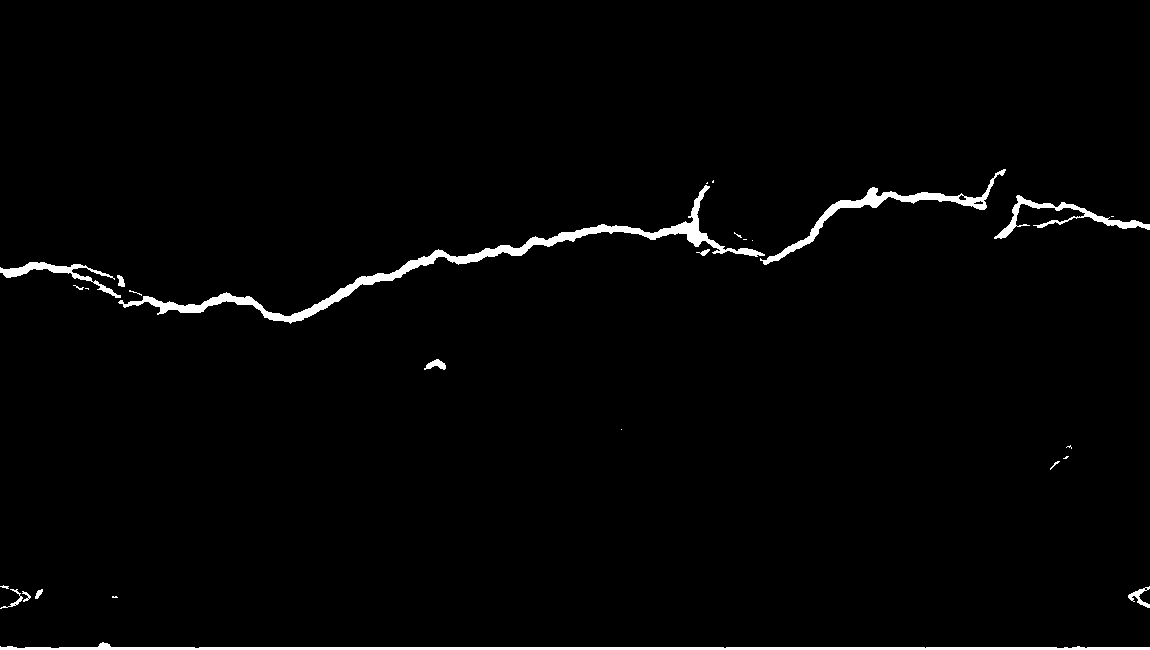}\\
    fine-tuned multiscale 3D U-Net from \ref{sec:unet-finetuned-multiscale} & \includegraphics[height=\myscaletwo]{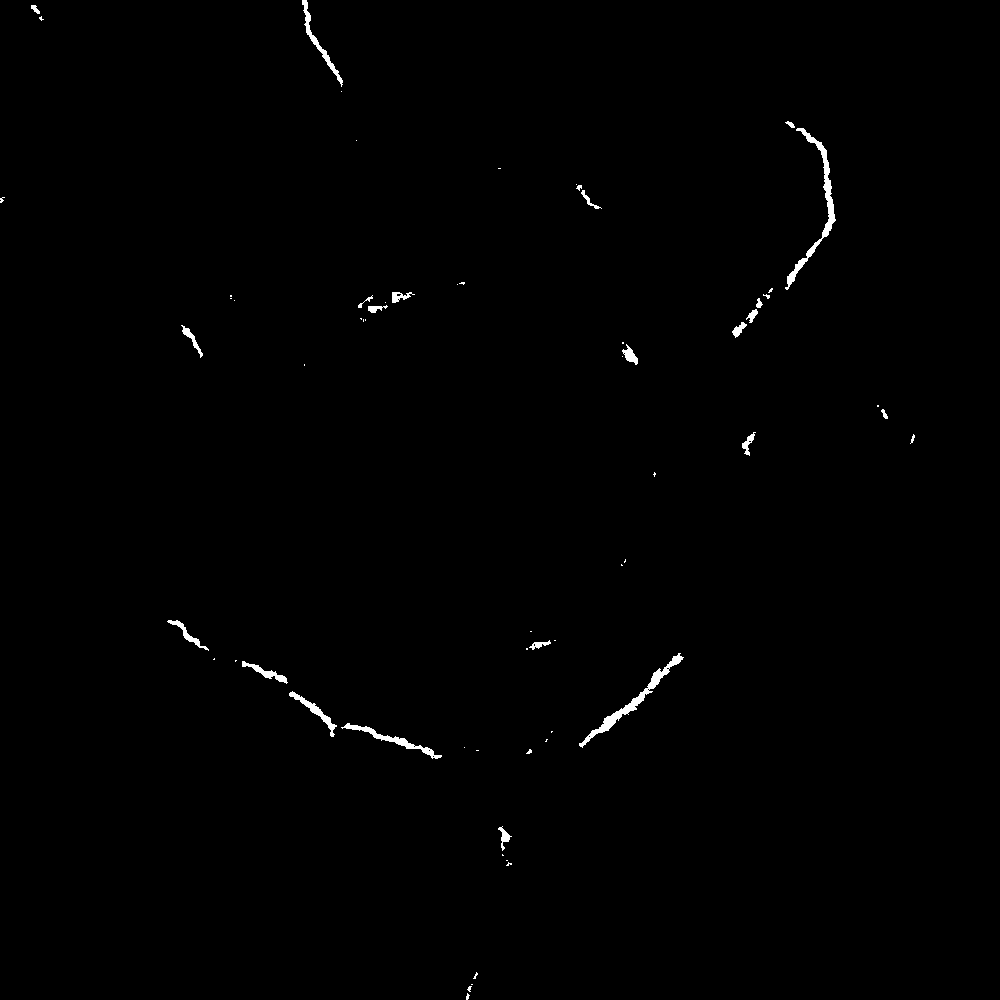} & \includegraphics[height=\myscaletwo]{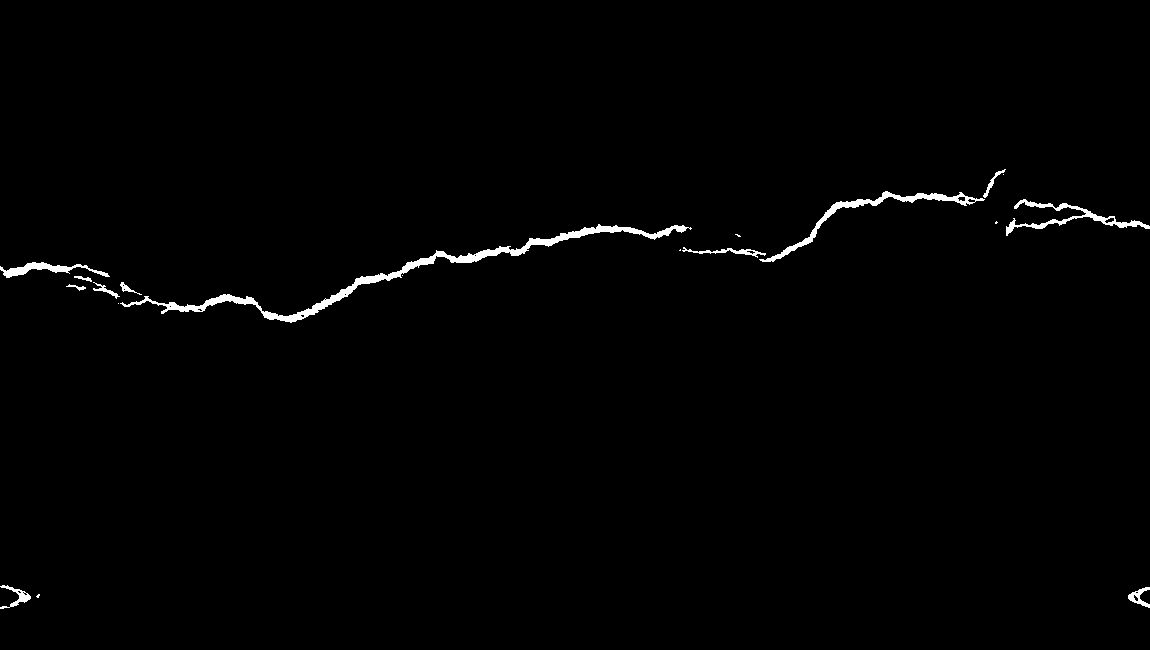}\\
    Riesz network as described in \ref{sec:riesz-network} & \includegraphics[height=\myscaletwo]{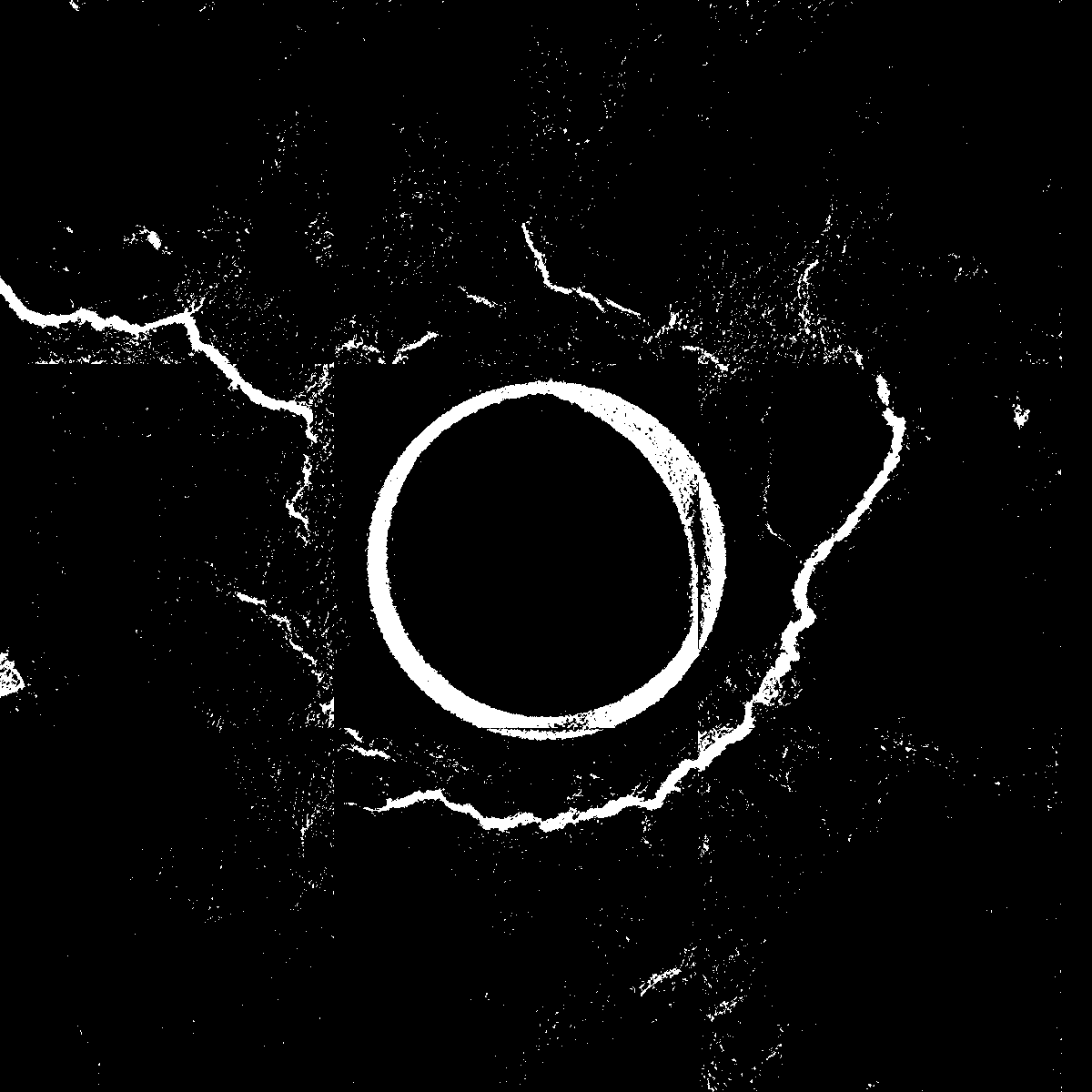} & \includegraphics[height=\myscaletwo]{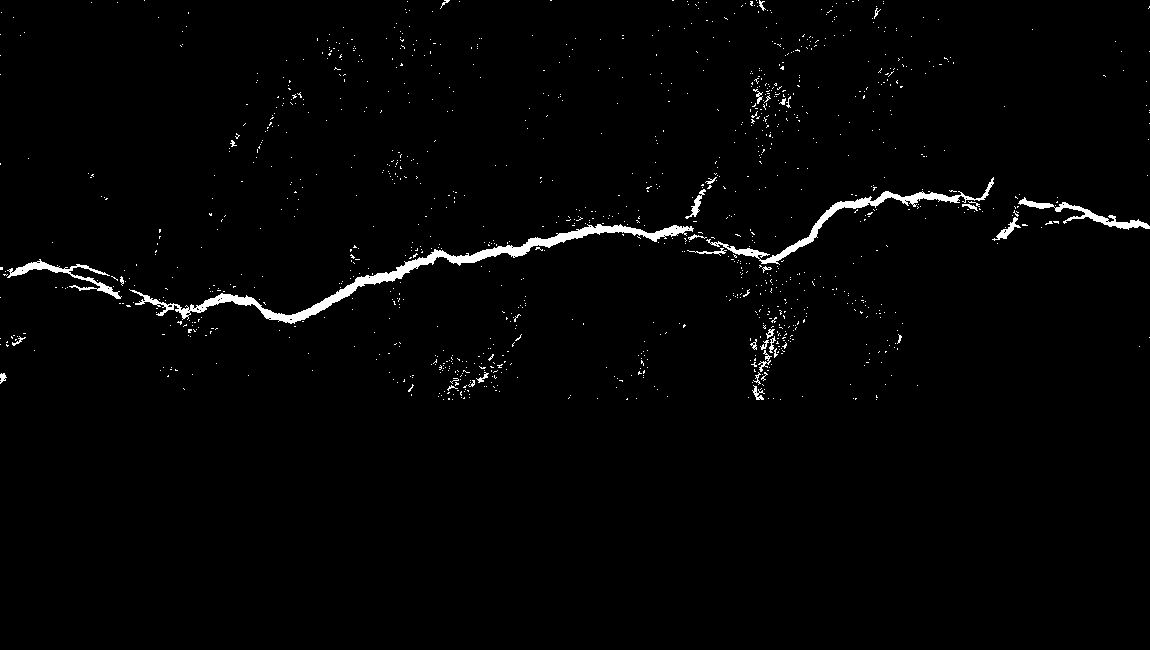}\\
    \svhline
    \end{tabular}
\end{table}

\section{Challenge 3: Fiber reinforced concrete}

Traditionally, the brittle material concrete is reinforced by thick 
steel rebars. Mixing much smaller fibers directly into the 
concrete matrix to improve mechanical strength and allow for freer design 
has a long history, too. Attempts to replace the rebars completely are 
more recent and subject of current research.
Examples of steel and polymer fiber reinforced concrete samples
are shown on the right side of Figure~\ref{fig:crack-examples-3d}. 

\begin{figure}[b]
%\sidecaption
\subfloat[SFRC, original]{\includegraphics[width=0.32\textwidth]{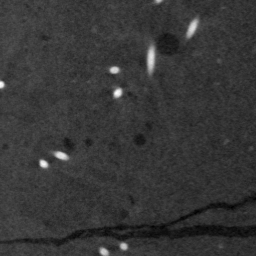}}
\hfill
\subfloat[3D U-Net]{\includegraphics[width=0.32\textwidth]{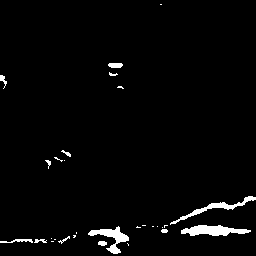}}
\hfill
\subfloat[3D U-Net, trained on FRC]{\includegraphics[width=0.32\textwidth]{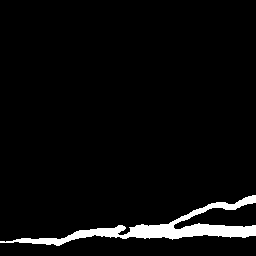}}\\
\subfloat[PPFRC, original]{\includegraphics[width=0.32\textwidth]{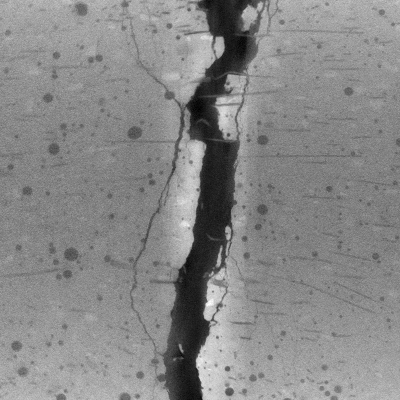}}
\hfill
\subfloat[3D U-Net]{\includegraphics[width=0.32\textwidth]{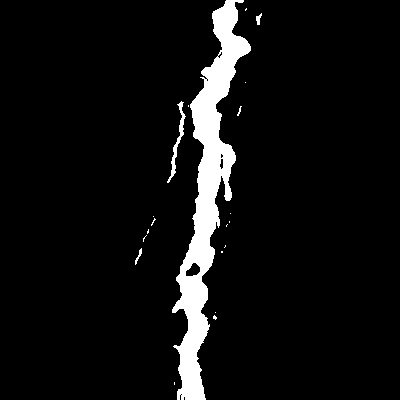}}
\hfill
\subfloat[3D U-Net, trained on FRC]{\includegraphics[width=0.32\textwidth]{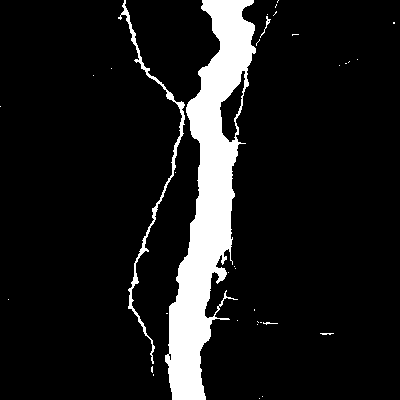}}\\
\caption{Examples of segmentation results for fiber reinforced concrete. 
The 3D U-Net not trained on FRC mistakes dark regions close to the 
bright steel fibers as cracks (b) and does not detect the thin branches 
of the crack in the PPFRC (e). The fine tuned 3D U-Net performs much 
better although classifying some pp fibers as crack (f).}
\label{fig:results-frc}      
\end{figure}

These fibers increase the complexity of our segmentation problem even 
further. Polymer fibers are thin, dark structures and thus easily 
mistaken to be cracks. Steel fibers absorb X-rays much stronger than 
the other microstructure components and thus require adapted imaging 
parameters which in turn result in more similar gray values for pores, 
cracks, and 
concrete matrix. Even worse, often, steel fibers feature a thin dark halo 
either due to having been coated by a resin enhancing bonding or due to imaging 
artifacts arising from the high local difference in absorption.
As a consequence, our 3D U-Nets trained on concrete without fibers fail, 
see the central column of Figure~\ref{fig:results-frc}.

In order to train a CNN robust with respect to reinforcing fibers, 
too, we generate synthetic crack images with the background 
acquired from CT images of fiber reinforced concrete samples. 
We train a 3D U-Net as described in Section~\ref{sec:cnn-for-3d-crack-segm}, 
now on $24$ images featuring one or two cracks of widths 
$1,\,3,$ and $5$ on backgrounds of normal and high performance concrete, 
as well as concretes reinforced by polypropylene and steel fibers. 
Detailed results are presented in \cite{nowacka23}. Here, we demonstrate 
the effect using two examples in Figure~\ref{fig:results-frc}.

\section{Discussion and conclusion}
Overall, is ML definitely able to solve our segmentation problem. This is however 
only possible thanks to the simulated training data. A wide variety of new research
topics arise from this. Most prominent is the question which properties of the 
synthetic training images enable generalization to their real world counterparts.
The Riesz network is a new, promising alternative to the abundant CNN. Already in 
the very experimental form presented here, it can compete with 3D U-Net. The 
strikingly low number of parameters that have to be fit during training renders it
very attractive in all cases where scale invariance is a natural requirement and 
annotated training data are scarce. 

For the practical problem to be solved, challenges remain, too. First of all, 
generalization to unseen concrete formulations. Does it just require additional 
training using the well known simulated cracks and one CT image of the 
new concrete in the spirit of a calibration? Or is it possible to train a model 
that finds cracks of all shapes in images of all types of concrete? In the 
latter case, how is correctness of this automatic generalization ensured?

Clearly, new opportunities open up once the cracks are correctly segmented. Geometric 
characterization of the crack structures help to better understand the material 
concrete as well as material systems involving concrete \cite{pullout-test-paper}. 
In-situ four-point bending tests within the XXL CT device Gulliver 
\cite{salamon:gulliver} will allow to observe crack initiation and evolution
in concrete beams of a length of more than $1\,$m. Analysis of the induced local
motion fields \cite{nogatz2022} may even unveil cracks too thin to be segmented 
based on just a static image. 
Finally, the geometric modelling can be refined, maybe in 
combination with multi-scale modelling of the concrete matrix. 

\begin{acknowledgement}
This work was supported by the German Federal Ministry of Education and 
Research (BMBF) [grant number 05M2020 (DAnoBi)]. 
We thank Matthias Pahn and Szymon Grezsiak from CE of RPTU for samples and
experimental design, Franz Schreiber from Fraunhofer ITWM and Michael Salamon 
from Fraunhofer EZRT for CT imaging, and Ali Moghiseh from Fraunhofer ITWM for 
input on CrackNet.
\end{acknowledgement}

\end{document}